\pdfoutput=1

\documentclass[11pt]{article}

\usepackage{EMNLP2023}

\usepackage{times}
\usepackage{latexsym}
\usepackage{graphicx}
\usepackage{tabularx}
\usepackage{booktabs}
\usepackage{adjustbox}
\usepackage{amsmath}
\usepackage{enumitem}
\usepackage[T1]{fontenc}

\usepackage[utf8]{inputenc}

\usepackage{microtype}

\usepackage{inconsolata}

\title{LogicPrpBank: A Corpus for Logical Implication and Equivalence}

\author{Zhexiong Liu$^{1}$\thanks{\;\;These authors contributed equally to this work.} , Jing Zhang$^{2}$\footnotemark[1] , Jiaying Lu$^{2}$, Wenjing Ma$^{2}$, Joyce C Ho$^{2}$
\\
$^{1}$University of Pittsburgh, Pittsburgh, Pennsylvania 15260 USA\\
$^{2}$Emory University, Atlanta, Georgia 30322 USA \\
   \texttt{zhexiong@cs.pitt.edu}, \texttt{jing.zhang2@emory.edu} \\ \texttt{\{jiaying.lu, wenjing.ma, joyce.c.ho\}@emory.edu}
  }

\begin{document}
\maketitle

\begin{abstract}
Logic reasoning has been critically needed in problem-solving and decision-making. Although Language Models (LMs) have demonstrated capabilities of handling multiple reasoning tasks (e.g., commonsense reasoning), their ability to reason complex mathematical problems, specifically propositional logic, remains largely underexplored. This lack of exploration can be attributed to the limited availability of annotated corpora. Here, we present a well-labeled propositional logic corpus, \textsc{LogicPrpBank}, containing 7093 Propositional Logic Statements (PLSs) across six mathematical subjects, to study a brand-new task of reasoning logical \textit{implication} and \textit{equivalence}. We benchmark \textsc{LogicPrpBank} with widely-used LMs to show that our corpus offers a useful resource for this challenging task and there is ample room for model improvement.
\end{abstract}

\section{Introduction}

Propositional logic deals with propositions (i.e., statements that can be true or false) and logical relationships between propositions. It has been used to solve many scientific problems (e.g., computer logic gates, distributed computing, and game strategies)~\citep{pietarinen2001propositional} and facilitate educational applications such as Intelligent Tutoring Systems (ITSs)~\citep{galafassi2020evologic,mandal2021classifying}. However, reasoning with propositional logic is different from reasoning in a Natural Language Processing (NLP) task (e.g., causal inference, commonsense reasoning, etc.) as propositional logic is a formal language formed with a set of symbols and rules that are distinct from those in natural languages~\citep{traylor-etal-2021-mean}. Table~\ref{tab:example} shows a truth table of logical \textit{implication} and \textit{equivalence} using a propositional theory for reasoning. As an illustration, the entailment of Propositional Logic Statement (PLS) P $\rightarrow$ Q\footnote{Note that statements P and Q do not need to be semantically related in propositional logic.}, \textit{``If the sum of the interior angles of a triangle is greater than 180 degrees, then a square has five sides''}, is a true statement given both P and Q are false, but these statements (P $\rightarrow$ Q, P, Q) are all incorrect from the commonsense perspective.

\begin{table}[t]
\centering
\begin{tabular}{cc|cc}
\toprule
P & Q & P $\rightarrow$ Q & P $\leftrightarrow$ Q \\ \midrule
T & T & T & T \\
T & F & F & F \\
F & T & T & F \\
F & F & T & T \\\bottomrule
\end{tabular}
\caption{The truth table of logical \textit{implication} (P $\rightarrow$ Q) and \textit{equivalence} (P $\leftrightarrow$ Q). Note that the logical \textit{implication} is always true given statement P is false, regardless of the truth value of statement Q, which is counterintuitive in commonsense.}
\label{tab:example}
\end{table}

Recently, Language Models (LMs) have demonstrated strong abilities to solve mathematical problems, e.g., approximating solutions to Partial Differential Equations~\citep{li2022transformer}, solving simple math word problems~\citep{patel2021nlp}, and reasoning arithmetic and logical problems~\citep{wang2021exploring}. It has been proven that increasing the scale of LMs (e.g., the size of model parameters) can lead to better performance and sample efficiency in many NLP tasks~\citep{devlin2018bert,brown2020language,kasneci2023chatgpt}. However, this claim is questionable in the propositional logic field because LMs are pre-trained with corpora that incorporate rationale in natural languages while reasoning with propositional logic requires understating rationale defined in formal languages. While \citet{traylor-etal-2021-mean} study a simple set of propositional logic (e.g., \textit{and}, \textit{or}, \textit{negation}) by investigating under which conditions LMs can successfully emulate the meaning of formal languages, this only reveals a small portion of propositional logic problems, but more complex ones (e.g., \textit{implication} and \textit{equivalence}) are underexplored. To bridge the gap, we present \textsc{LogicPrpBank} that contains 7093 PLSs in six mathematical subjects (\textit{algebra, arithmetic, calculus, geometry, number theory (numbers),} and \textit{statistics}) to investigate the LMs' capabilities of reasoning complex propositional logic. This endeavor is beneficial to mathematical ITSs.

There have been available corpora for evaluating LMs' abilities to understand and reason. \citet{bowman2015large} present the \textsc{SNLI} corpus that focuses on single-step inferences (e.g., entailment, contradiction, irrelevance) between two pieces of text. \textsc{SNLI} is unable to explain reasoning chains, although an extension has been later implemented in~\citet{camburu2018snli}.  \citet{ribeiro2022entailment} introduce \textsc{EntailmentBank} corpus that investigates the entailment relations of natural language text. However, these datasets focus on propositional logic inference (e.g., entailing conclusions from premises) but not the fundamental correctness of PLSs in mathematical subjects. Recently, \citet{ontanon2022logicinference} collected a propositional logic corpus \textsc{LogicInference}, which validates the subset of first-order propositional logics using sequence-to-sequence LMs. In comparison to \textsc{LogicInference} which focuses on reasoning between premises and conclusions, \textsc{LogicPrpBank} has two major differences. First, we focus on mathematical PLSs which can be used for building educational applications (e.g., ITSs). Second, instead of dealing with logical inference (e.g., inference chains), \textsc{LogicPrpBank} investigates the correctness of PLSs (e.g., the truth values of PLSs). 

With \textsc{LogicPrpBank}, our work investigates two research questions: \textbf{RQ1} Are LMs capable of reasoning complex propositional logic (e.g., \textit{implicating} and \textit{equivalence}) in real mathematical subjects? \textbf{RQ2} Are large-scale LMs better than small-scale LMs in reasoning propositional logic? We benchmark \textsc{LogicPrpBank} and make the following contributions:

\begin{itemize}[leftmargin=*,itemsep=-1.8pt, topsep=1.5pt]
  \item We leverage the state-of-the-art ChatGPT to generate real atomic PLSs in six mathematical subjects and then develop a proposition composer to compose atomic to compound PLSs.
  \item We investigate LM's capability of reasoning complex \textit{implication} and \textit{equivalence} PLSs which is different from reasoning in existing NLP tasks.
  \item We conduct experiments on LogicPrpBank with various scales of LMs to study the pros and cons of LMs in reasoning with propositional logic.
\end{itemize}

\section{Corpus}

\begin{table*}[t]
\centering
\begin{adjustbox}{width=1\textwidth}
\begin{tabular}{ccclc}
\toprule
\textbf{IDs} & \textbf{Types} & \textbf{Subjects} & \textbf{Propositional Logic Statements} & \textbf{Truth Values} \\ \midrule
1 & Atomic & Arithmetic & The median of 3, 4, 5, 6, 7 is 6. & F \\  \midrule
2 & Atomic & Geometry & The distance between two parallel lines is the same at all points. & T \\ \midrule
3 & Atomic & Numbers & The sum of the first n odd integers is n(n+1). & F \\ \midrule
4 & Implication & Geometry & \begin{tabular}[l]{@{}l@{}} The distance between two parallel lines is the same at all points is necessary \\ and sufficient for the area of a circle is always pi * r * r, where r is the diameter. \end{tabular} & F \\ 
\midrule
5 & Equivalence & Calculus & \begin{tabular}[l]{@{}l@{}} The derivative of log(x) with respect to x is equal to 1/x is equivalent \\ to the implicit function theorem only applies to functions of two variables.\end{tabular} & T\\ 
\bottomrule
\end{tabular}
\end{adjustbox}
\caption{The examples of the proposed \textsc{LogicPrpBank} corpus.}
\label{tab1}
\vspace{-2.8mm}
\end{table*}

We use ChatGPT API\footnote{\url{https://openai.com/blog/chatgpt}} to generate atomic PLSs. In particular, we develop a data collection prompt to collect True and False PLSs: \texttt{please list [X] [Y] atomic statements in [Z]}, where \texttt{[X]} is the number of PLSs (e.g., \texttt{X}=20), \texttt{[Y]} is chosen from \textit{True} or \textit{False}, and \texttt{[Z]} is chosen from one of the subjects (\textit{algebra, arithmetic, calculus, geometry, number theory,} and \textit{statistics}). We run the same prompts multiple times until having substantial True and False PLSs for each subject.

We use ChatGPT as a corpus source rather than open sources (e.g., online articles) or human annotations for two reasons. First, ChatGPT is trained with vast amounts of data from the internet written by humans, which covers propositional logic lectures across educational and tutoring webpages, thus it is able to generate a large number of high-quality PLSs. Also, using ChatGPT to generate a corpus is a new exploration of LMs' applications in corpus construction which minimizes labor costs associated with collection and reduces annotation costs. Second, understanding mathematical PLSs necessitates annotators acquire mathematical knowledge at the college or even higher education levels, thus it is not feasible for annotators to create a large number of PLSs from scratch. Therefore, we conduct a pilot exploration of corpus construction using ChatGPT. Note that a ChatGPT-generated PLS contains two-dimensional information: one is the statement itself; another is the truth value of the statement. To validate the correctness (true/false) of ChatGPT-generated PLSs, we employ qualified human annotators who pass a qualification test to check the ChatGPT-generated PLSs. Annotators are asked to check whether or not a ChatGPT-generated PLS is matched to its truth values by using annotators' expert knowledge, checking online resources, or referencing textbooks. Each ChatGPT-generated PLS is checked by one annotator. We observe that the ChatGPT-generated PLSs have a 17.4\% error rate, where a ChatGPT-generated false PLS is proved to be True; and vice versa. We then ask annotators to manually correct the wrong statements by revising the PLSs to match their correct truth values. We randomly sample 10\% atomic PLSs from each subject and ask two annotators to annotate their truth values without seeing ChatGPT-generated truth values. The Cohen's kappa between the two annotators is 0.77.

To generate \textit{implication} and \textit{equivalence} PLSs, we develop a template-based proposition composer with curated templates to automatically compose two atomic PLSs into one compound PLS. An \textit{implication} composer uses a set of templates: (1) \texttt{if [P] then [Q]}; (2) \texttt{[P] implies [Q]}; (3) \texttt{[P], therefore, [Q]}. An \textit{equivalence} composer uses a set of templates: (1) \texttt{[P] if and only if [Q]}; (2) \texttt{[P] is necessary and sufficient for [Q]}; (3) \texttt{[P] is equivalent to [Q]}. Here \texttt{[P]} and \texttt{[Q]} denote two different atomic PLSs. Accordingly, the labels (i.e., truth values) of compound PLSs are inferred from their truth table (see Table~\ref{tab:example}, where \textit{implication} is the column of P $\rightarrow$ Q and \textit{equivalence} is the column of P $\leftrightarrow$ Q). We collect 1277 atomic PLSs that cover axioms, theorems, and practice problems. We randomly sample one P and one Q from the same subject to generate compound PLSs (P $\rightarrow$ Q and P$\leftrightarrow$ Q) by running the proposition composers. After several rounds of the process, 5816 compound PLSs are generated. Table~\ref{tab1} shows examples from the \textsc{LogicPrpBank} corpus, where a compound PLS is composed of two atomic PLSs (e.g., the \textit{implication} PLS in Row\#4 uses the atomic PLS in Row\#2). Table \ref{tab2} shows the statistics of atomic, \textit{implication}, and \textit{equivalence} PLSs regarding their truth values (true/false) across six mathematical subjects. Note that \textit{implication}  has more true PLSs than false PLSs because P$\rightarrow$Q is always true given P is false, regardless of the value of Q (see Table~\ref{tab:example}).
The true/false ratio in atom and \textit{equivalence} is near one.

\begin{table*}[tbp]
\centering
\begin{adjustbox}{width=0.9\textwidth}
\begin{tabular}{c|cccccc|c}
\toprule
\textbf{Types} & \textbf{Algebra} & \textbf{Arithmetic} & \textbf{Calculus} & \textbf{Geometry} & \textbf{Numbers} & \textbf{Statistics} & \textbf{Total} \\ \midrule
Atomic & 115 / 132 & 117 / 117 & 101 / 110 & 115 / 117 & 43 / 60 & 122 / 128 & 613 / 664 \\
Implication & 466 / 144 & 410 / 138 & 338 / 107 & 405 / 133 & 83 / 23 & 465 / 160 & 2167 / 705 \\
Equvilance & 338 / 287 & 295 / 266 & 236 / 220 & 299 / 253 & 53 / 56 & 363 / 278 & 1584 / 1360 \\ \midrule
Total & 919 / 563 & 822 / 521 & 675 / 437 & 819 / 503 & 179 / 139 & 950 / 566 & 4364 / 2729 \\
\bottomrule
\end{tabular}
\end{adjustbox}
\caption{The statistics of \textsc{LogicPrpBank} corpus across subjects, atom, \textit{implication}, and \textit{equivalence} PLSs. The \# before/after slash is the \# of True and False PLSs, respectively.}
\label{tab2}
\end{table*}

\section{Experiments and Analysis}

In this section, we introduce the benchmark experiments on \textsc{LogicPrpBank} corpus for PLS correctness (true/false) classification. We use small-scale LMs, e.g., DistilRoBERTa~\citep{sanh2019distilbert}, RoBERTa-base~\citep{liu2019roberta}, BERT-base and BERT-large~\citep{devlin2018bert}, medium-scale LMs, e.g., GPT2-medium~\citep{radford2019language}, and BLOOM-560m~\citep{scao2022bloom}, and large-scale Language Models (LLMs), e.g., Llama2-7B~\citep{touvron2023llama},   to reason atomic, \textit{implication}, and \textit{equivalence} PLSs, respectively. 
 We finetune small- and medium-scale LMs with parameter sizes ranging from dozens million to half a billion (given limited computing resources), and perform few-shot learning on LLMs.
In particular, we evaluate the performance of Llama2-7B on the test set with zero-shot, 1-shot, 3-shot, 5-shot, and 10-shot learning. In the zero-shot scenario, we predict the test set results without using any training examples. In the other few-shot experiments, we retrieve top-n examples from the training set that are most similar to the test example as context. The similarity is determined by cosine function between sentence embedding~\citep{reimers2019sentence}.
We split the corpus into the train (70\%), validation (10\%), and test (20\%) sets. We use the training set to train small- and medium-scale LMs and use the validation set to tune their parameters.
In the training, we optimize the model using Adam optimizer with $\beta_1=0.9, \beta_2=0.999, \epsilon=10^{-8}$. The learning rate is $10^{-5}$ and the batch size is 32. We train 20 epochs and select the best model on the validation.  We conduct three-seed runs and report average macro F1 scores on the test set for both finetuning and zero-shot/few-shot learning. We implement LMs with PyTorch~\citep{paszke2019pytorch} and initial model weights from HuggingFace \footnote{\url{https://huggingface.co}}. Due to space limitation, we introduce prompt details in our code repository\footnote{\url{https://github.com/JZCS2018/AI4ED2024}}. The experiments are running on a GeForce RTX 3090 GPU. 

\begin{figure}[t]
    \centering
    \includegraphics[width=0.5\textwidth]{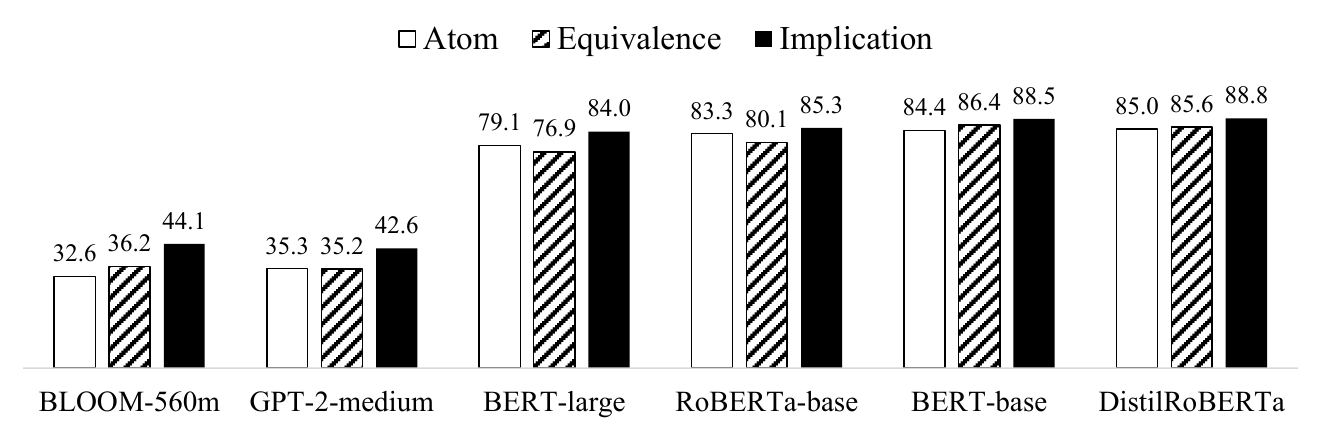}
    \caption{LM performance on \textsc{LogicPrpBank} across atom, \textit{implication}, and \textit{equivalence} PLS.}
    \label{fig:bar_chat}
\end{figure}

Table \ref{tab:3} shows the results of zero-shot/few-shot and finetuned LMs for identifying the correctness of PLSs in six mathematical subjects. We have observed that small-scale LMs perform excellently in \textit{calculus}, \textit{geometry}, and \textit{statistics} but are dramatically poor in \textit{arithmetic} and \textit{number theory}. This might be because \textit{arithmetic} and \textit{number theory} have many number-related propositions (see Table~\ref{tab1}) that are known deficiencies of LMs. These observations answer \textbf{RQ1} that LMs are able to reason complex propositional logic only on specific mathematical subjects. Figure~\ref{fig:bar_chat} shows the results of finetuned small- and medium-scale LMs and pre-trained LLMs on atomic, \textit{equivalence}, and \textit{implication}. We observe that \textit{implication} is generally better than the other two. BERT and RoBERTa (LMs) have good performance but medium-scale LMs (BLOOM-560m and GPT-2-medium) and LLMs have poor performance. These observations answer \textbf{RQ2} that small-scale LMs are able to reason complex propositional logic but Llama2 fails, which suggests that increasing the size of LMs results in performance degradation (see BERT-base v.s. BLOOM-560m). Although large-scale LMs are supposed to have better performance and sample efficiency in many downstream NLP tasks, they do not hold true in reasoning with propositional logic. We argue that propositional logic is a formal language that uses different logic theories from those in natural languages (e.g., commonsense knowledge). Therefore, the medium-scale LMs might not learn propositional logic well given the limited size of the corpus for training. The LLMs are slightly better than the medium-scale LMs, which suggests that LLMs may have potential in learning propositional logics, even in a few-shot learning scenario. Moreover, we observe that the 5-shot learning yields the most favorable overall performance, whereas there is an unexpected decline in performance with the 10-shot configuration. This implies that it is possible that more examples provided to LLMs would introduce more noise. In conclusion, the constructed \textsc{LogicPrpBank} is helpful for training small-scale LMs to learn complex propositional logic reasoning in most subjects.

\begin{table*}[tbp]
\centering

\begin{adjustbox}{width=0.98\textwidth}

\begin{tabular}{c|c|cccccc|c} \toprule
\textbf{Models}                       & \textbf{Sizes} & \textbf{Algebra}         & \textbf{Arithmetic}      & \textbf{Calculus}        & \textbf{Geometry}        & \textbf{Numbers}         & \textbf{Statistics}      & \textbf{Overall}          \\ \midrule
\multicolumn{1}{c|}{Llama2-zeroshot}  & 7B             & 39.07       & 38.37       & 34.68       & 38.53       & 36.17       & 40.11       & 38.28        \\
\multicolumn{1}{c|}{Llama2-1shot}   & 7B             & 41.19       & 38.37       & 44.43       & 42.17       & 40.26       & 43.89       & 42.09        \\
\multicolumn{1}{c|}{Llama2-3shot} & 7B             & 45.70       & 42.29       & 43.94       & 45.10       & 46.67       & 46.92       & 45.08        \\
\multicolumn{1}{c|}{Llama2-5shot}  & 7B             & 53.30       & 39.09       & 54.29       & 47.69       & 38.44       & 48.61       & 46.94        \\
\multicolumn{1}{c|}{Llama2-10shot}   & 7B             & 43.72       & 39.88       & 43.29       & 45.52       & 35.48       & 45.39       & 43.59        \\ \midrule
BLOOM-560m                            & 560M           & 39.80          & 37.96          & 38.23          & 39.82          & 36.23          & 39.19          & 38.98           \\
GPT-2-medium                          & 345M           & 39.08          & 38.52          & 37.82          & 39.34          & 36.74          & 38.67          & 38.69           \\ \midrule
BERT-large                            & 340M           & 77.83          & 49.28          & 96.40          & 92.70          & 52.85          & 97.03          & 81.03           \\
RoBERTa-base                          & 125M           & 83.57          & 51.71          & 96.80          & 95.53          & 44.22          & 96.15          & 83.36           \\
BERT-base                             & 110M           & \textbf{92.00} & \textbf{56.71} & \textbf{98.94} & \textbf{98.19} & \textbf{66.93} & \textbf{99.30} & \textbf{87.65}  \\
DistilRoBERTa                         & 82M            & 91.54          & 55.78          & 98.77          & 97.59          & 56.67          & 98.87          & 87.27           \\ \bottomrule
\end{tabular}

\end{adjustbox}
\caption{The F1 scores of zeroshot/fewshot and finetuned LMs on \textsc{LogicPrpBank} across six subjects. The top rows are LLMs, the middle rows are medium-scale LMs, and the bottom rows are small-scale LMs. The highest F1 scores are in bold.}
\label{tab:3}
\end{table*}

\section{Related Work}
Previous research has focused on addressing mathematical problems within the field of NLP \citep{wang-etal-2017-deep,saxton2019analysing,dua-etal-2019-drop}. These studies utilize a question-answering framework to tackle these mathematical problems. The Math23L corpus \citep{wang-etal-2017-deep} consists of basic English contextual information, equations, and corresponding answers, primarily involving arithmetic problems. DeepMind's research \citep{saxton2019analysing} investigates reasoning processes in algebra. The DROP corpus \citep{dua-etal-2019-drop} is a reading comprehension corpus that includes various types of mathematical tasks, such as subtraction and selection. Notably, all answers to its questions can be directly or indirectly inferred from the provided passages. These questions bear similarities to those found in Math23L, and the corpus is sourced from elementary school math word problems. Based on their limitations, we propose \textsc{LogicPrpBank} corpus to address PLSs in six subjects, which poses a brand-new task of reasoning propositional logic.

\section{Conclusion}

Reasoning propositional logic (e.g., \textit{implication} and \textit{equivalence}) differs from reasoning in NLP tasks (e.g., commonsense reasoning), since the former involves adhering to a set of rules expressed in formal languages, while the latter requires understanding about the real world and the common knowledge conveyed in natural language. To date, there are limited corpora and studies focusing on propositional logic, especially in mathematical subjects. To bridge the gap, we present \textsc{LogicPrpBank}, a corpus containing 7093 atom, \textit{implication}, and \textit{equivalence} proposition statements, designed to facilitate LMs to reason complex mathematical propositional logic. The experiments indicate that LLMs (e.g., Llama2-7B) and medium-LMs (e.g., BLOOM-560m) perform worse than the lighter and faster LMs (e.g., BERT and RoBERTa), and LMs struggle with reasoning \textit{arithmetic} and \textit{number theory} but promising in \textit{calculus}, \textit{geometry}, and \textit{statistics}. In future work, we plan to extend the \textsc{LogicPrpBank} corpus to encompass a wider range of subjects, including physics and chemistry, in order to support the development of interdisciplinary ITSs.

\section*{Limitations}
Our corpus was collected from ChatGPT and then verified and/or annotated by a qualified annotator, thus there could be annotating errors that would influence the accuracy of the experiments. Using ChatGPT to generate a corpus is a brand-new design, which brings challenges to verify and validate the quality of the data, e.g., the inter-agreement rate used in a traditional data annotation pipeline. But our designed verify-then-correct process for data collecting is proven to save time and labor. Also, ChatGPT is not free for the whole community so it would not be available to researchers from specific areas, thus using ChatGPT to generate or annotate data needs extra ethical considerations. In addition, our corpus is relatively small, which makes it difficult to train or even finetune LLMs. Although zeroshot/fewshot learning with Llama2 shows promising results but still not better than trained/finetuned small-scale LMs, which suggests that LLMs are not always the best options while solving specific tasks that have small annotated data. Furthermore, we only focus on propositional logic in mathematical fields, however, logical reasoning is not limited to math subjects but to many real-world scenarios that we have not covered. Moreover, we only benchmark the corpus on a small number of LMs due to computational resource limitations. And, the error analysis regarding the performance of LMs on our proposed corpus is not extensively studied because most LMs are difficult to visualize and/or explain their reasoning steps (e.g., Llama2, CPT-2, BLOOM, etc.). Therefore, we have limited discussions about error analysis in this work. 

\section*{Acknowledgments}
The research was supported by the National Science Foundation Award IIS-2145411. The opinions expressed were those of the authors and did not represent the views of the institutes. We would like to thank anonymous reviewers for their valuable feedback on this work.
\bibliography{anthology,custom}
\bibliographystyle{acl_natbib}

\end{document}